%% file: main.tex
\documentclass[conference]{IEEEtran}
\usepackage{times}

\usepackage[numbers]{natbib}
\usepackage{multicol}
\usepackage[bookmarks=true]{hyperref}

\usepackage{amssymb}
\usepackage{booktabs}
\usepackage{multirow}
\usepackage{wrapfig}
\usepackage{graphicx}
\usepackage{amsmath} 
\usepackage{algorithm}
\usepackage{algorithmicx}
\usepackage{algpseudocode}  % or use algorithmic
\usepackage{lipsum}  
\usepackage{float}
\usepackage{caption}
\title{XPG-RL: Reinforcement Learning with Explainable Priority Guidance for Efficiency-Boosted Mechanical Search}

\author{
  Yiting Zhang$^{1}$\textsuperscript{\textdagger},
  Shichen Li$^{2}$, 
  and Elena Shrestha$^{1}$ \\
  $^{1}$University of Michigan, Ann Arbor\hspace{5pt} $^{2}$University of Illinois, Urbana-Champaign\\
}

\begin{document}

% \begin{figure}[H]
%     \centering
%     \includegraphics[width=0.9\textwidth]{figures/demo.pdf}
%     \caption{We propose XPG-RL, a reinforcement learning framework for mechanical search tasks. XPG-RL leverages task-guided action prioritization and learns context-aware switching over action primitives, effectively reducing redundant manipulations and improving task efficiency. The figure shows the manipulator successfully grasping a target object (\textit{banana}) in a densely cluttered real-world scene.}
%     \label{fig:demo}
%     \vspace{-0.2cm}
%   \end{figure}

\makeatletter
    \let\@oldmaketitle\@maketitle%
    \renewcommand{\@maketitle}{
    \@oldmaketitle
    \centering
    \vspace{-2.5pt}
    \includegraphics[width=0.9\textwidth]{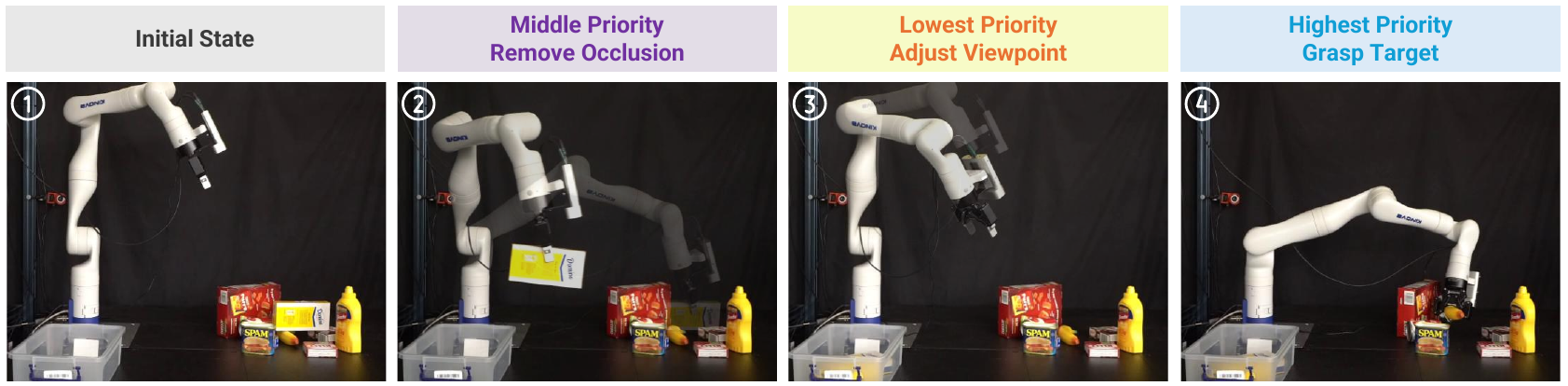}
    \captionof{figure}{
    We propose XPG-RL, a reinforcement learning framework for mechanical search tasks. XPG-RL leverages task-guided action prioritization and learns context-aware switching over action primitives, effectively reducing redundant manipulations and improving task efficiency. The figure shows the manipulator successfully grasping a target object (\textit{banana}) in a densely cluttered real-world scene.}
    \label{fig:demo}
    \vspace{-12pt}
    \addtocounter{figure}{-1}%
    }
\makeatother

% avoiding spaces at the end of the author lines is not a problem with
% conference papers because we don't use \thanks or \IEEEmembership

%Homer Simpson\authorrefmark{2},
%Montgomery Scott\authorrefmark{3} and
%Eldon Tyrell\authorrefmark{4}}
%Telephone: (800) 555--1212, Fax: (888) 555--1212}
%\authorblockA{\authorrefmark{4}Tyrell Inc., 123 Replicant Street, Los Angeles, California 90210--4321}}

\maketitle

\begingroup
\renewcommand\thefootnote{\textdagger}
\footnotetext{Corresponding author. Email: yitzhang@umich.edu.}
\endgroup

\addtocounter{figure}{1}

\begin{abstract}
Mechanical search (MS) in cluttered environments remains a significant challenge for autonomous manipulators, requiring long-horizon planning and robust state estimation under occlusions and partial observability. In this work, we introduce XPG-RL, a reinforcement learning framework that enables agents to efficiently perform MS tasks through explainable, priority-guided decision-making based on raw sensory inputs. XPG-RL integrates a task-driven action prioritization mechanism with a learned context-aware switching strategy that dynamically selects from a discrete set of action primitives such as target grasping, occlusion removal, and viewpoint adjustment. Within this strategy, a policy is optimized to output adaptive threshold values that govern the discrete selection among action primitives. The perception module fuses RGB-D inputs with semantic and geometric features to produce a structured scene representation for downstream decision-making. Extensive experiments in both simulation and real-world settings demonstrate that XPG-RL consistently outperforms baseline methods in task success rates and motion efficiency, achieving up to 4.5$\times$ higher efficiency in long-horizon tasks. These results underscore the benefits of integrating domain knowledge with learnable decision-making policies for robust and efficient robotic manipulation. The project page for XPG-RL is \href{https://yitingzhang1997.github.io/xpgrl/}{https://yitingzhang1997.github.io/xpgrl/}.
\end{abstract}

\IEEEpeerreviewmaketitle

\input{sections/01_introduction}
\input{sections/02_related_works}
\input{sections/03_method}
\input{sections/04_experiments}
\input{sections/05_conclusion}

\bibliographystyle{plainnat}
\bibliography{references}

\clearpage

\appendix
\input{appendix/training_details}
\input{appendix/nbv}
\input{appendix/additional_exp}

\end{document}

%% file: sections/01_introduction.tex
\section{Introduction}
\label{sec:introduction}
Mechanical search (MS) is a critical yet challenging task in robotic manipulation. It requires robots to intelligently plan and execute sequences of physical interactions to locate and retrieve occluded objects from cluttered environments~\cite{danielczuk2019mechanical}. Unlike traditional pick-and-place tasks, MS involves occlusion, long-horizon planning, and environmental uncertainty, where robots must infer object locations, plan sequential actions, and adapt to unpredictable dynamics~\cite{li2020towards, nam2021fast}. These challenges hinder real-world applications such as logistics automation, warehouse management, and domestic assistance~\cite{d2023cluttered, kurenkov2021semantic}, highlighting the need for more efficient and reliable MS methods.

Given the dynamic and unpredictable nature of MS tasks, predefining all possible scenarios is impractical. Learning-based approaches are essential for allowing robots to adapt to new situations and generalize beyond predefined rules~\cite{chen2024differentiable, zhang2025certifiably}. Reinforcement learning (RL) enables robots to learn policies directly from visual inputs through trial-and-error exploration, without requiring explicit object pose knowledge, and is widely recognized as an effective approach for tackling MS tasks~\cite{zeng2018learning, kurenkov2020visuomotor, liu2023synergistic, wang2024multi}. However, conventional RL methods treat each action primitive independently via separate neural networks, neglecting action interdependencies~\cite{imtiaz2023prehensile, tang2023selective}. This limitation leads to cascading errors and can misguide decision-making in scenarios requiring long-horizon planning. Zenkri et al. observed that when the number of objects increased from 10 to 20, the efficiency of traditional RL methods dropped by 46.6\%~\cite{zenkri2022hierarchical}. These challenges highlight the need for advancements in RL to enhance long-term MS performance and improve decision-making efficiency.

To enhance interactions between actions in long-horizon planning, recent studies have explored hierarchical reinforcement learning (HRL) for complex MS tasks~\cite{zenkri2022hierarchical, wang2021i2hrl}. HRL introduces an additional network on top of standard RL structures for high-level action selection, enabling the learning of action interdependencies across sequences~\cite{wang2023hierarchical}. However, MS training typically relies on limited datasets, which constraints HRL’s capacity to learn effectively. In short-term tasks, this limitation often result in over-analyzing procedural steps due to insufficient data coverage. In more complex, long-horizon tasks, HRL further struggles with embedding and feature extraction, as learning abstract high-level representations implicitly from sparse data becomes increasingly challenging. Research has shown that HRL often lacks the granularity to capture subtle environmental variations~\cite{schiewer2024exploring}, leading to suboptimal action selection, inefficient exploration, and compounding errors over extended task horizons. Thus, HRL remains limited in data-inefficient MS tasks.

The primary challenge in improving MS efficiency lies in fully utilizing high-dimensional, limited, and hard-to-distinguish data~\cite{kroemer2021review}. Addressing this requires advancements in both model design and data enrichment. On one hand, enhancing model architectures can reduce the complexity of learning tasks. By alleviating the training burden from raw, complex inputs, models can better focus on essential features, improving task success rates~\cite{meng2024meta}. In MS tasks, incorporating predefined guidance into the model design can streamline learning by directing attention to critical information. On the other hand, enriching the dataset is important. In robot learning, a key difficulty is capturing relevant targets during in-situ image collection. Instead of relying solely on fixed viewpoints and actions within a limited knowledge scope, increasing information sources can significantly improve efficiency by providing more comprehensive contextual understanding~\cite{li2025multi}.

% In this paper, we propose XPG-RL (\textbf{R}einforcement \textbf{L}earning with e\textbf{X}plainable \textbf{P}riority \textbf{G}uidance), a novel framework for MS problems. 
The main contributions of this work are four-fold:

\begin{itemize}
    \item We propose a task-guided action prioritization mechanism that embeds domain knowledge into a structured hierarchy over manipulation primitives. This promotes modular, explainable high-level policy learning and reduces learning complexity in long-horizon tasks.
    \item We present XPG-RL, a novel RL framework that leverages the prioritization structure to learn context-aware switching between action primitives. This approach minimizes redundant or misaligned actions and improves overall task efficiency.
    \item We integrate a next-best-view (NBV) strategy, enabling the robot to actively select informative viewpoints. This improves perception in occluded scenes and allows for spatial understanding beyond what is possible from a fixed viewpoint.
    \item We validate our approach in both simulated and real-world cluttered environments, demonstrating that XPG-RL consistently outperforms strong baselines in task success rate and motion efficiency.
\end{itemize}

%% file: sections/02_related_works.tex
\section{Related Works}
\label{sec:related_works}

\textbf{Visual-input-only MS.} 
Recent research has explored RL for directly learning MS policies from visual inputs, without relying on explicit object pose estimation. One of the most influential frameworks in this space is Visual Pushing Grasping (VPG)~\cite{zeng2018learning}, which learns separate quality maps for pushing and grasping primitives and selects actions by choosing the pixel with the highest predicted quality. Numerous follow-up works have refined this paradigm by improving network architectures~\cite{kumra2022learning, imtiaz2023prehensile, tang2023selective}. To improve high-level decision-making, HRL has been explored to better capture the interactions among action primitives, especially in long-horizon tasks~\cite{nasiriany2022augmenting, strudel2020learning, lee2022hierarchical, yang2021hierarchical}. Our work is inspired by the HRL structure while focused on the high-level decision-making.

\textbf{Priority-guided decision-making in robot manipulation.}
In robot manipulation, integrating task or action priorities has long been a common strategy for managing complex decision-making under constraints. Traditional methods typically rely on fixed priority rules or cost-based heuristics to sequence actions~\cite{salvietti2009task, kaelbling2011hierarchical, dogar2012planning}. More recent work has introduced learning-based frameworks that incorporate human priors or domain-specific logic to guide high-level decisions~\cite{zeng2021transporter, li2024mpgnet}. However, these approaches still depend on manually designed, experience-driven heuristics at the high level, which limits their adaptability. In long-horizon or highly cluttered tasks, where scene configurations can vary widely, such static logic often fails to generalize or respond effectively to unseen situations.

%% file: sections/03_method.tex
\section{Method}
\label{sec:method}

\begin{figure*}[t]
    \centering
    \includegraphics[width=0.85\textwidth]{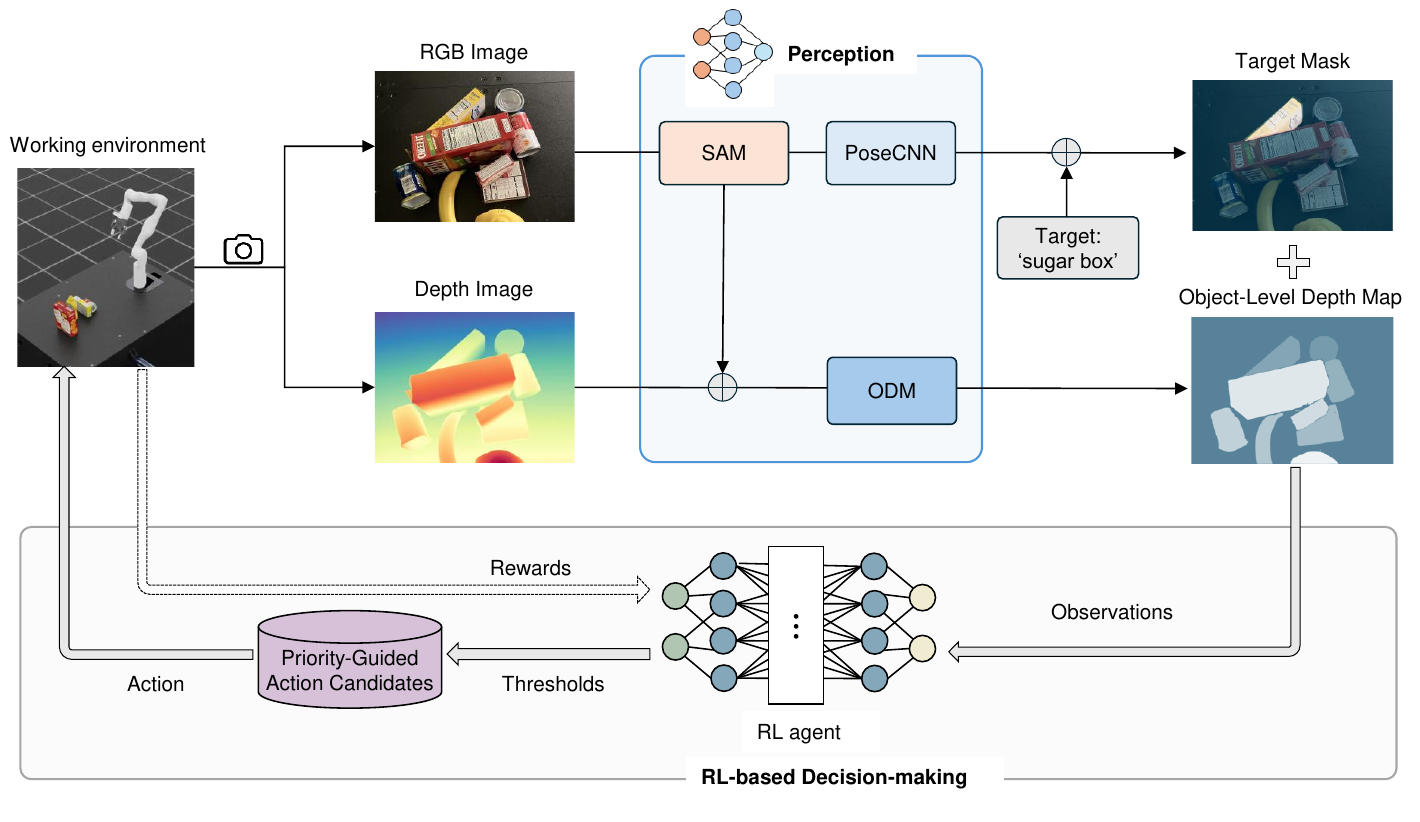}
    \caption{\textbf{Overview of the XPG-RL framework.} The framework consists of two main components: (1) a perception pipeline, which processes fused RGB-D images to extract semantic and geometric context and builds a compact scene representation; and (2) an RL-based decision-making module, which takes this representation as input and learns a policy to predict adaptive thresholds. These thresholds guide the selection among priority-structured action candidates (e.g., target grasping, occlusion removal, and viewpoint adjustment), enabling context-aware and efficient action execution.
    }
    \label{fig:method_overview}
    \vspace{-0.5cm}
\end{figure*}

Our MS task formulation is based on a framework ~\cite{zenkri2022hierarchical}, where the objective is to retrieve an occluded target object from a cluttered scene using a manipulator equipped with a parallel-jaw gripper. The robot perceives the scene through an eye-in-hand RGB-D camera, but the initial viewpoint is often contrained by occlusions. To address this challenge, we propose XPG-RL (\textbf{R}einforcement \textbf{L}earning with e\textbf{X}plainable \textbf{P}riority \textbf{G}uidance). The method comprises two main components: (1) a perception pipeline and, (2) an RL-based decision-making module (Fig.~\ref{fig:method_overview}).

% An overview of our proposed XPG-RL method is illustrated in Fig.\ref{fig:method_overview}. At each step, the RGB-D camera captures the current scene, producing RGB and depth images. These are processed by a perception pipeline composed of the Segment Anything Model (SAM)\cite{kirillov2023segment}, PoseCNN~\cite{xiang2017posecnn}, and an Object Depth Mapping (ODM) module. SAM and PoseCNN collaborate to generate an accurate mask for the specified target object (e.g., the sugar box), while ODM fuses the masks of all objects in the scene with depth data to produce an object-level depth map. These outputs form the observation input to the RL agent. The agent then selects the most promising manipulation action by integrating this observation with priority-guided (PG) candidate actions. Upon executing the selected action, the system transitions to a new state and returns an immediate reward as feedback for policy learning.

\subsection{Perception}

The perception pipeline receives fused RGB-D images captured from the working scene. The RGB image provides visual appearance cues, such as surface property and material type. We leverage the Segment Anything Model (SAM)~\cite{kirillov2023segment} to generate appearance-based masks for all objects, and integrate it with a PoseCNN~\cite{xiang2018posecnn} model to add semantic context, producing a mask focused on the specified target object.

The depth image captures spatial geometry useful for understanding occlusion, volume, and spatial relationships. However, raw depth data often varies across different surfaces of the same object, making it sub-optimal for object-level reasoning. To address this, we propose an Object Depth Mapping (ODM) module to generate a simplified, object-centric spatial representation. Specifically, the raw depth image is fused with object masks from the RGB image. Instead of relying on pixel-wise depth, we compute the average depth for each object. Objects are then sorted by distance, from farthest to nearest, and rendered in progressively lighter shades to generate the final object-level depth map. The target-specific mask and the object-level depth map are jointly provided as input to the RL agent, enabling efficient environmental interpretation and informed decision-making.

\subsection{RL-based Decision-making}

We formulate the decision-making process as a partially observable Markov decision process (POMDP). At each time step $t$, the perception pipeline outputs an observation $o_{t}$, which provides partial information on the true underlying environment state $s_{t}$. The agent follows a policy $\pi(\tau_{t} \mid o_{t})$, which maps the observation $o_t$ to a set of decision thresholds $\tau_{t} = (\tau_{1, t}, \tau_{2, t})$, where $\tau_{1, t}, \tau_{2, t} \in [0, 1]$. To inform action selection, the system predicts quality scores $Q_{\text{target}, t}$ and $Q_{\text{occlude}, t}$ using AnyGrasp~\cite{fang2023anygrasp}. representing the estimated success probabilities of grasping the target object and removing an occluding object, respectively. These thresholds $\tau_{t}$ are applied to the predicted scores to determine which action $a_{t} \in \mathcal{A} = \left\{ a_{1}, a_{2}, ..., a_{n} \right\}$ to execute from a set of priority-guided action candidates, such as grasping the target or removing an occluding object. The priority-guided action selection process is detailed in Algorithm~\ref{alg:grasp_planning}. After executing the selected action, the environment transitions to a new state $s_{t+1}$, and the agent receives an immediate reward $r_{t}$.
% The agent's decision-making is guided by policy $\pi(\tau_{t} \mid o_{t})$, which maps the observations to a set of thresholds $\tau_{t} = (\tau_{1, t}, \tau_{2, t})$ with $\tau_{1, t}, \tau_{2, t} \in [0, 1]$. These thresholds are used to select an action $a_{t} \in \mathcal{A} = \left\{ a_{1}, a_{2}, ..., a_{n} \right\}$, which controls switching between a set of priority-guided (PG) action candidates such as grasping the target or removing an occluding object. The quality scores $Q_{\text{target}, t}$, $Q_{\text{occlude}, t}$ are predicted using AnyGrasp~\cite{fang2023anygrasp} and represent the estimated success probabilities of grasping the target object and removing the occluder, respectively. Based on the selected thresholds and predicted scores, one of the actions $a_{t} \in \mathcal{A}$ is selected and executed. The system then transitions to a new state $s_{t+1}$ and the agent receives an immediate reward $r_{t}$ . 
The reward function encourages successful grasps while penalizing unnecessary or infeasible actions, and is defined as the following:
\begin{equation}
r_t =
\begin{cases}
1000, & \text{if the target is successfully extracted;} \\
-100, & \text{if an infeasible action is selected;} \\
-1, & \text{otherwise.}
\end{cases}
\end{equation}

The goal is to maximize the expected cumulative discounted reward $R_{t} = \sum_{i=0}^\infty \gamma^{i} r_{t+i}$
% \begin{equation}
%     R_{t} = \sum_{i=0}^\infty \gamma^{i} r_{t+i},
% \end{equation}
where $\gamma \in [0, 1]$ is the discount factor that balances the importance of future versus immediate rewards. 
% The overall decision process is summarized in Algorithm~\ref{alg:grasp_planning}.

% \begin{wrapfigure}{r}{0.5\textwidth}
% \vspace{-1.5em}  % adjust spacing as needed
% \begin{minipage}{0.5\textwidth}
% \begin{algorithm}[H]
% \caption{Priority-Guided Decision Policy}
% \label{alg:grasp_planning}
% \begin{algorithmic}
% \Require Thresholds \( \tau_t = (\tau_{1,t}, \tau_{2,t}) \)
% \State Compute grasp score \( Q_{\text{target}, t} \)
% \If{ \( Q_{\text{target}, t} > \tau_{1,t} \) }
%     \State Grasp the target object
% \Else
%     \State Compute \( Q_{\text{occlude}, t} \)
%     \If{ \( Q_{\text{occlude}, t} > \tau_{2,t} \) }
%         \State Grasp and remove an occluding object
%     \Else
%         \State Move camera to new viewpoint
%     \EndIf
% \EndIf
% \end{algorithmic}
% \end{algorithm}
% \end{minipage}
% \vspace{-1em}  % tweak bottom spacing
% \end{wrapfigure}

\begin{algorithm}[H]
\caption{Priority-Guided Decision Policy}
\label{alg:grasp_planning}
\begin{algorithmic}
\Require Thresholds \( \tau_t = (\tau_{1,t}, \tau_{2,t}) \)
\State Compute grasp score \( Q_{\text{target}, t} \)
\If{ \( Q_{\text{target}, t} > \tau_{1,t} \) }
    \State Grasp the target object
\Else
    \State Compute \( Q_{\text{occlude}, t} \)
    \If{ \( Q_{\text{occlude}, t} > \tau_{2,t} \) }
        \State Grasp and remove an occluding object
    \Else
        \State Move camera to new viewpoint
    \EndIf
\EndIf
\end{algorithmic}
\end{algorithm}
\vspace{-0.5cm}

To achieve this objective, we employ Proximal Policy Optimization (PPO)~\cite{schulman2017proximal} to optimize the policy $\pi(\tau_{t} \mid o_{t})$, which determines continuous thresholds $\tau_{t}$ that guide action selection. This approach allows for flexible decision boundaries over predicted success probabilities, supporting effective learning under partial observability.

\begin{figure*}[!t]
    \centering
    \includegraphics[width=0.8\textwidth]{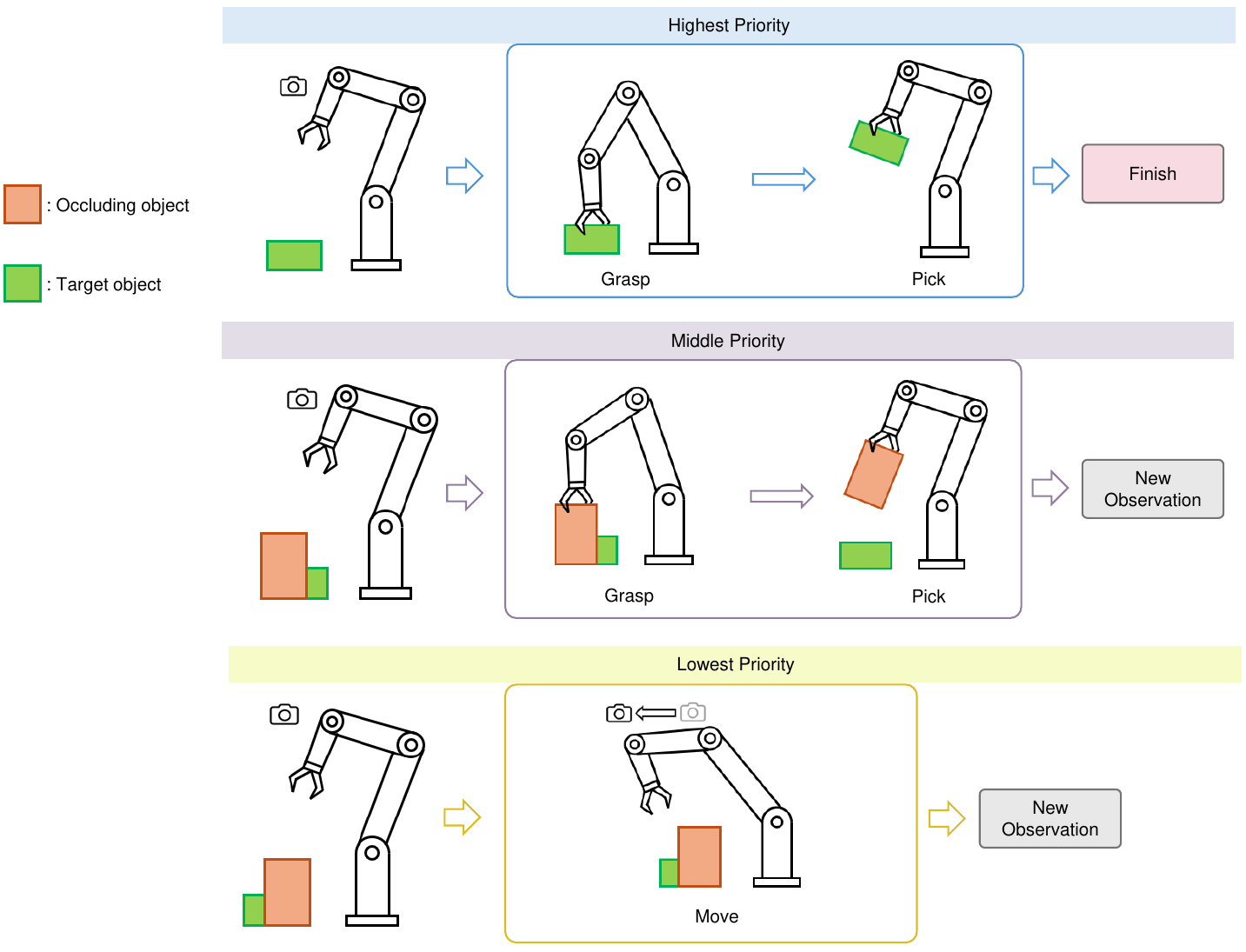}
    \caption{\textbf{Priority-guided action candidates.} The action space consists of three discrete primitives—target grasping, occlusion removal, and viewpoint adjustment—ranked from highest to lowest priority. Learned thresholds govern switching between these actions, enabling the agent to make efficient, interpretable decisions by sequentially evaluating actions in priority order.}
    \label{fig:actions}
    \vspace{-0.5cm}
\end{figure*}

Furthermore, to enhance sample efficiency and promote robust behavior, we constrain the agent’s action space using a domain-informed set of priority-guided candidates (Fig.~\ref{fig:actions}). These structured discrete actions, ranked by priority, simplify the decision-making process and ensure reliable performance, even with limited training data. For the MS task, we propose the following three priority-guided action candidates:
% The agent uses the threshold values $\tau_{t}$ along with the predicted quality scores $Q_{\text{target}, t}$ and $Q_{\text{occlude}, t}$ to follow predefined PG action candidates. These candidates are human-defined discrete actions (Fig~\ref{fig:actions}) ranked by priority, which constrain the action space and support more sample-efficient learning and reliable behavior classification, even with limited training data. We propose the following three PG action candidates for the MS task: 

\begin{itemize}
    \item \textbf{Highest priority}: If the grasp quality for the target object is high, i.e., \( Q_{\text{target}, t} \ge \tau_{1, t} \), the agent directly executes the grasp with the highest quality score on the target object. This action is the final execution of one MS task. 
    \item \textbf{Middle priority}: If the target grasp quality is low but the quality for occluding objects is high, i.e., \( Q_{\text{target}, t} < \tau_{1, t} \) and \( Q_{\text{occlude}, t} \ge \tau_{2, t} \), the agent selects the occluding object with the highest score for removal. After the action is executed, the agent re-evaluates whether the condition for the highest-priority action is now satisfied.
    \item \textbf{Lowest priority}: If both the target and occluding objects have low grasp quality, i.e., \( Q_{\text{target}, t} < \tau_{1, t} \) and \( Q_{\text{occlude}, t} < \tau_{2, t} \), the agent initiates a next-best-view (NBV) strategy where it generates a set of novel candidate viewpoints. It constructs a Truncated Signed Distance Function (TSDF) representation of the current scene, simulates each candidate view, predicts the grasp quality for the target, and selects the viewpoint with the highest predicted quality as the next camera pose. 
    % This NBV strategy, inspired by an affordance-based NBV approach~\cite{zhang2023affordance}, enriches data collection by increasing viewpoint diversity, providing more comprehensive scene understanding and improving generalization across tasks. Unlike prior work, our method selectively activates NBV only when no higher-priority actions are viable, allowing the agent to avoid unnecessary manipulation and instead focus on gathering informative observations to enhance task efficiency.
\end{itemize}

%% file: sections/04_experiments.tex
\section{Experiments}
\label{sec:experiments}

We evaluate the performance of XPG-RL in both simulation and real-world environments. We use a 7-DoF Kinova robot arm equipped with a RealSense camera and a parallel jaw ROBOTIQ gripper at the end effector. The objects used in the experiments are selected from the YCB object set~\cite{calli2015benchmarking}. We train the RL policy using the PPO algorithm as implemented in Stable-Baselines3~\cite{raffin2021stable}. Further details on the training hyperparameters and network architecture are provided in the appendix.
% The simulation platform is built upon Isaac Lab~\cite{mittal2023orbit} due to its compatibility with our working environment setup and its realistic dynamic modeling. The hardware platform is built upon ROS 2~\cite{macenski2022robot} due to its high scalability and adaptivity to various robotic applications.

\textbf{Metrics.} In our experiments, we evaluate the performance of our method using the following metrics:
\textbf{Task success rate} is defined as the percentage of episodes in which the target object is successfully grasped and lifted at least 15\,cm above the table within 10 motions.
\textbf{Average motion number} refers to the mean quantity of actions needed to achieve the task per episode. 
\textbf{Task efficiency} is defined as the ratio between task success rate and average motion number, reflecting how effectively a method achieves successful outcomes using minimal actions.

\textbf{Baselines.} We compare the performance of our method with the following baselines. \textbf{Target-Oriented VPG} is a traditional RL method that extends the original VPG~\cite{zeng2018learning} by incorporating the target mask as an additional input. It outputs quality maps for pushing and grasping, executing the action with the highest quality score. \textbf{Hierarchical Policy Learning}~\cite{zenkri2022hierarchical} is an HRL approach that uses a high-level policy network on top of traditional RL architectures to model action interdependencies across sequences. \textbf{MPGNet}~\cite{li2024mpgnet} builds upon Target-Oriented VPG by jointly learning three actions: (1) moving, (2) pushing, and (3) grasping. It adopts a hierarchical structure and incorporates a simple heuristic-based logic for high-level action selection.

\begin{table*}[!t]
\centering
\caption{\textbf{Success rate and efficiency across scenes.} XPG-RL consistently achieves higher success rates with fewer actions than baselines, demonstrating robust and efficient performance.}
\resizebox{0.8\textwidth}{!}{
\begin{tabular}{lcccccccccc}
\toprule
\multirow{2}{*}{\textbf{Method}} & \multicolumn{4}{c}{\textbf{Task success rate [\%]}} & \multicolumn{4}{c}{\textbf{Average motion number}} \\
\cmidrule(lr){2-5} \cmidrule(lr){6-9}
& 5 objects & 10 objects & 15 objects & 20 objects & 5 objects & 10 objects & 15 objects & 20 objects \\
\midrule
Target-Oriented VPG & 60 & 30 & 26 & 15 & 4.92 & 5.38 & 7.85 & 8.01 \\
Hierarchical Policy Learning & 57 & 53 & 30 & 21 & 3.91 & \underline{4.06} & \underline{5.18} & \underline{7.45} \\
MPGNet & \underline{76} & \underline{59} & \underline{35} & \underline{22} & \underline{3.84} & 4.51 & 6.78 & 7.65 \\
XPG-RL & \textbf{81} & \textbf{76} & \textbf{71} & \textbf{64} & \textbf{2.31} & \textbf{2.77} & \textbf{3.60} & \textbf{4.94} \\
\bottomrule
\end{tabular}
}
\label{tab:sim_comp_with_baseline}
\vspace{-1em}
\end{table*}

\subsection{Simulation Experiments}

The task is simulated in Isaac Lab~\cite{mittal2023orbit}, which provides realistic dynamic modeling capabilities. We generate batches of simulated scenes with varying levels of complexity. Each batch consists of 100 scenes populated with $n$ objects randomly selected from the YCB object set and distributed on a table within a 0.5 m $\times$ 0.5 m square area.

\subsubsection{Comparison to Baselines}
% \vspace{-0.1cm}
We evaluate the performance of XPG-RL against several baseline methods across a range of scene complexities, defined by the number of objects in the environment: from 5 (simple cases) to 20 (highly cluttered cases). As object count increases, the scene becomes more occluded and challenging. Table~\ref{tab:sim_comp_with_baseline} summarizes the task success rate and average motion number for each method.

\begin{figure}[ht]
  \centering
  \includegraphics[width=0.45\textwidth]{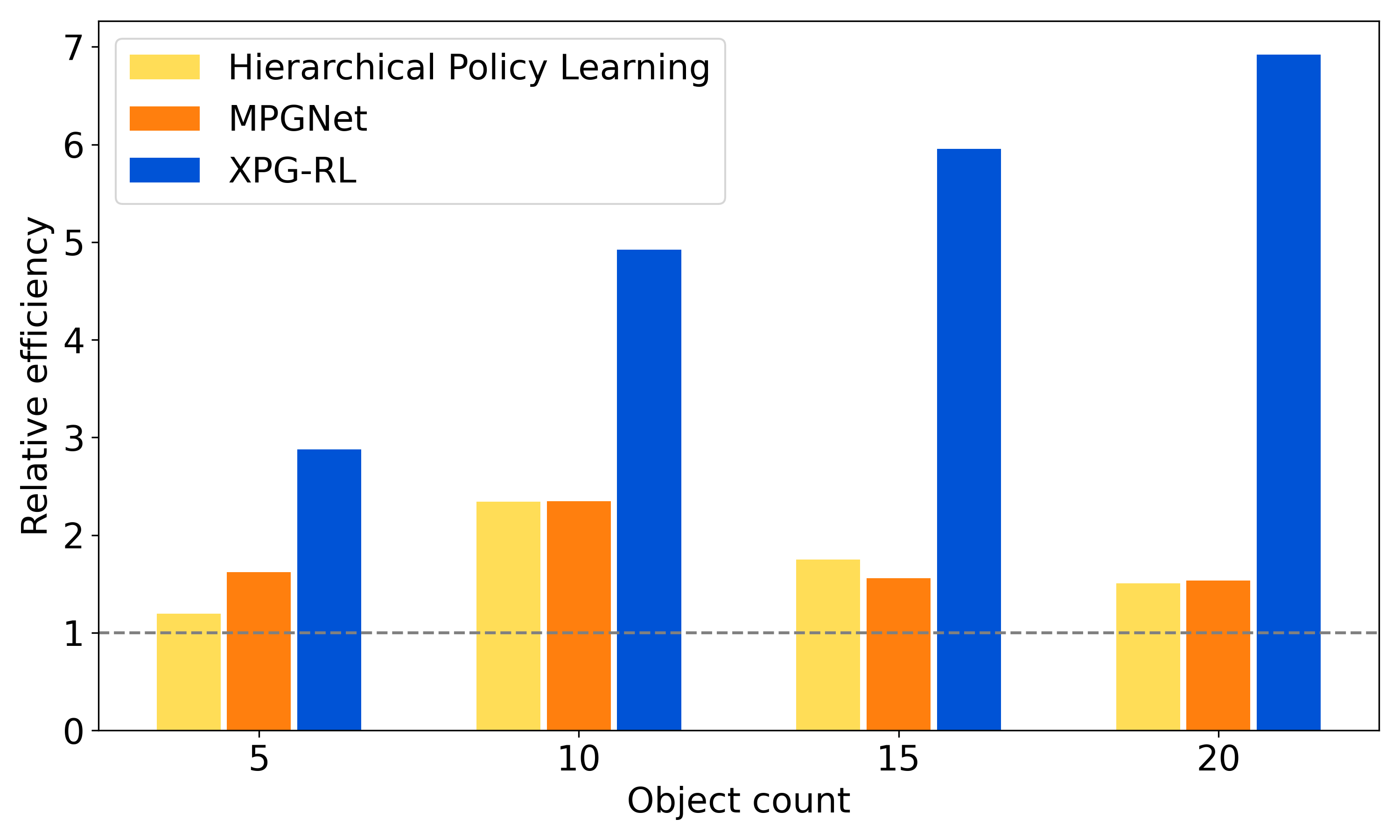}
  \caption{\textbf{Efficiency scaling with complexity.} XPG-RL outperforms baselines across all clutter levels, with relative efficiency gains widening as object count increases.}
  \label{fig:exp_efficiency}
  \vspace{-1em}
\end{figure}

Across all methods, we observe a general trend: the task success rate decreases and the average motion number increases as scene complexity grows. Some notable patterns emerge. In relatively simple scenarios (e.g., 5 objects), Hierarchical Policy Learning tends to perform the worst. This is due to its multi-level decision-making structure, which can trigger unnecessary actions instead of a direct grasp. Target-Oriented VPG also struggles as complexity increases, highlighting its limited ability to coordinate multiple action primitives effectively. MPGNet benefits from its heuristic-driven high-level selection and performs reasonably well across general cases. However, its predefined logic lacks adaptability—it cannot dynamically adjust behavior based on task simplicity, nor flexibly respond to increasing scene complexity. As a result, it often triggers unnecessary actions in simple tasks and struggles with efficiency as complexity grows.

We further evaluate task efficiency by computing the relative efficiency of each method, using Target-Oriented VPG as the reference baseline. As shown in Figure~\ref{fig:exp_efficiency}, XPG-RL consistently demonstrates the highest efficiency across all scenarios. Notably, the efficiency gap between XPG-RL and the second-best method widens as episode length increases. In the 10-object case, XPG-RL achieves a relative efficiency of 5.0, while the baselines remain below 2.5. In the 20-object case—the most complex scenario—XPG-RL reaches 6.9, compared to 1.5 for both MPGNet and Hierarchical Policy Learning, yielding over a 4.5$\times$ improvement. This trend highlights XPG-RL’s stability and effectiveness in long-horizon tasks, whereas other methods tend to degrade in performance as complexity grows.

% \begin{figure}[ht]
%     \centering
%     \includegraphics[width=0.6\textwidth]{figures/exp_efficiency.png}
%     \caption{Relative task efficiency of different methods across increasing scene complexity, normalized to Target-Oriented VPG.}
%     \label{fig:exp_efficiency}
% \end{figure}

To further investigate whether XPG-RL retains its advantage in short-horizon tasks, we conduct an additional evaluation on scenes with 3 to 5 objects. The results are presented in Table~\ref{tab:sim_comp_with_baseline_simple} in the appendix. While both XPG-RL and MPGNet perform reasonably well in terms of task success, XPG-RL completes tasks with significantly fewer motions. This efficiency gain further underscores the benefit of learning-based high-level control, which enables XPG-RL to adapt its behavior based on task complexity and avoid redundant actions when a simple solution suffices.

These results highlight the effectiveness of XPG-RL’s design, which combines structured action prioritization with a learned high-level decision-making policy. By integrating a neural network to make context-aware decisions over a prioritized set of action primitives, XPG-RL achieves robust and efficient performance across both short- and long-horizon tasks, consistently outperforming all baseline methods.

\begin{table*}[!t]
\centering
\caption{\textbf{Contribution of priority guidance and NBV.} Removing priority guidance or NBV leads to lower success rates and efficiency, especially as scene complexity increases. This confirms that both components are critical for robust performance in cluttered environments.}
\resizebox{0.8\textwidth}{!}{
\begin{tabular}{lcccccccccc}
\toprule
\multirow{2}{*}{\textbf{Method}} & \multicolumn{4}{c}{\textbf{Task success rate [\%]}} & \multicolumn{4}{c}{\textbf{Average motion number}} \\
\cmidrule(lr){2-5} \cmidrule(lr){6-9}
& 5 objects & 10 objects & 15 objects & 20 objects & 5 objects & 10 objects & 15 objects & 20 objects \\
\midrule
No Priority Guidance & 64 & 57 & 52 & 42 & 4.87 & 5.28 & 5.75 & 6.24 \\
No NBV Motion & 78 & 65 & 62 & 51 & 2.27 & 2.87 & 3.27 & 3.93 \\
XPG-RL & 81 & 76 & 71 & 64 & 2.31 & 2.77 & 3.60 & 4.94 \\
\bottomrule
\end{tabular}
}
\label{tab:sim_ablation}
% \vspace{-1em}
\end{table*}

\begin{table*}[!t]
\centering
\caption{\textbf{Real-world performance across cluttered scenes.} XPG-RL consistently achieves higher task success rates with fewer actions than baselines across four tabletop environments.}
\resizebox{0.8\textwidth}{!}{
\begin{tabular}{lcccccccccc}
\toprule
\multirow{2}{*}{\textbf{Method}} & \multicolumn{4}{c}{\textbf{Task success rate}} & \multicolumn{4}{c}{\textbf{Average motion number}} \\
\cmidrule(lr){2-5} \cmidrule(lr){6-9}
& scene 1 & scene 2 & scene 3 & scene 4 & scene 1 & scene 2 & scene 3 & scene 4 \\
\midrule
Hierarchical Policy Learning & 2/5 & 3/5 & 3/5 & 0/5 & 6.4 & 3.8 & 5.0 & 5.8 \\
MPGNet & 3/5 & 4/5 & 3/5 & 2/5 & 3.4 & 4.8 & 4.4 & 5.2 \\
XPG-RL & 4/5 & 5/5 & 5/5 & 4/5 & 2.8 & 2.0 & 3.2 & 3.4 \\
\bottomrule
\end{tabular}
}
\label{tab:rw_comp_with_baseline}
\vspace{-0.5cm}
\end{table*}

\subsubsection{Ablation Study}
% \vspace{-0.1cm}

To assess the contribution of each key component in XPG-RL, we evaluate two ablated variants: (1) \textbf{No Priority Guidance} and (2) \textbf{No NBV Motion}. In the \textbf{No Priority Guidance} variant, the RL agent no longer uses the action priority scheme. Instead of learning threshold-based switching decisions among primitives, it directly learns a flat policy to select actions. The \textbf{No NBV Motion} variant retains the priority-guided decision-making policy but disables the NBV motion. If neither the target object nor the occluding object has a feasible grasp, the trial is considered a failure. This configuration isolates the importance of NBV in improving visibility and grasp success in occluded scenes.

The results, summarized in Table~\ref{tab:sim_ablation}, highlight the contribution of each module. Removing the priority guidance leads to a clear and consistent drop in both task success rate and motion efficiency, especially as scene complexity increases. This demonstrates that the structured action prioritization significantly improves decision quality, particularly in longer-horizon tasks. When comparing the full XPG-RL model to the No NBV Motion variant, we observe that NBV selection plays a relatively minor role in simpler scenes (e.g., 5 objects) but becomes increasingly important in complex and highly occluded environments. In such cases, NBV helps the agent improve visibility and locate graspable views of the target object, resulting in higher success rates and fewer failed attempts. Overall, the ablation study confirms that both priority guidance and NBV selection are essential for achieving robust, efficient performance, with their contributions becoming more pronounced as task complexity increases.

\subsection{Real-World Experiments}

\begin{figure}[ht]
    \centering
    \includegraphics[width=0.55\textwidth]{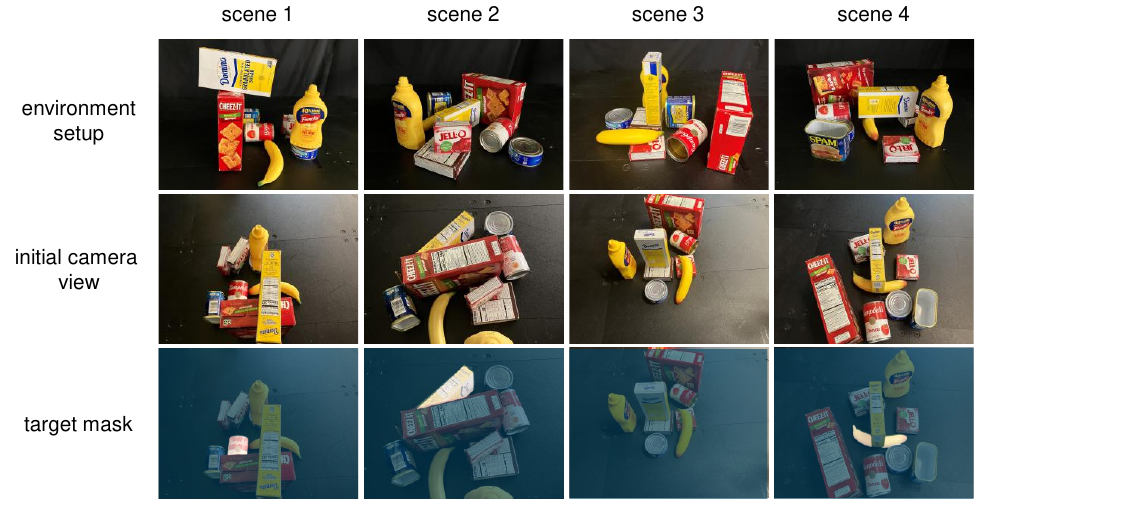}
    \caption{\textbf{Real-world experiment scenes.} Four tabletop setups with 9 objects each, illustrating different occlusion conditions, including a fully occluded target in Scene 3.}
    \label{fig:exp_rw_setup}
    \vspace{-0.5cm}
\end{figure}

We further evaluate the effectiveness and robustness of XPG-RL in real-world cluttered environments across four different tabletop scenes, each containing 9 randomly selected objects that are densely arranged on the table. The manipulator is positioned such that the target object is either fully or partially occluded in the initial camera view. To account for variability in initial object placement and manipulator positioning, each scene is repeated 5 times. The environment setup, initial view, and target object mask for each scene are shown in Fig.~\ref{fig:exp_rw_setup}. Notably, in Scene 3, the target object (\textit{meat can}) is completely occluded from the initial camera view, and thus no target pixels are visible in the mask.

The results are summarized in Table~\ref{tab:rw_comp_with_baseline}. Across all scenes, XPG-RL consistently achieves higher task success rates and requires fewer motions compared to baseline methods. In Scene 3, where the target object is fully occluded, XPG-RL effectively leverages NBV motions to reposition the camera, enabling successful identification and grasping. Rather than disturbing surrounding objects and worsening occlusion, it strategically selects a side viewpoint to minimize risk and maintain task feasibility. In Scene 4, the target object (\textit{banana}) is partially occluded by a \textit{sugar box}, where naive pushing or failed grasps could destabilize the scene and obscure the target further. XPG-RL avoids such failures through priority-guided decision-making that defers risky actions when predicted success probabilities fall below learned thresholds. This mechanism enables the agent to prioritize safer alternatives, such as NBV motions, and supports more informed, interpretable decisions. By learning context-aware thresholds, XPG-RL dynamically balances risk and reward based on the scene context, demonstrating strong generalization and reliable performance in real-world cluttered environments. 

%% file: sections/05_conclusion.tex
\section{Conclusion}
\label{sec:conclusion}

% Integrating domain knowledge with learnable decision-making policies offers a promising pathway toward robust and efficient robotic manipulation in mechanical search tasks. 

We propose XPG-RL, a reinforcement learning framework that enhances efficiency and robustness in mechanical search through structured action prioritization and adaptive action switching. XPG-RL consistently outperformed state-of-the-art methods in both simulation and real-world environments. In simulated cluttered scenes, XPG-RL demonstrated up to a 4.5$\times$ improvement in relative efficiency. In real-world experiments involving fully occluded scenes, XPG-RL completed all five trials with fewer actions, whereas state-of-the-art methods frequently failed. These results underscore the benefits of integrating domain knowledge with learnable decision-making policies for robust and efficient robotic manipulation.

%% file: appendix/training_details.tex
\subsection{Training Details}
\label{app:training_details}

\subsubsection{Policy Training Hyperparameters}
The RL agent is trained for a total of 10,000 steps using the PPO algorithm. The policy is optimized with a learning rate of $3\times10^{-4}$, a discount factor ($\gamma$) of 0.99, and a generalized advantage estimation (GAE) parameter ($\lambda$) of 0.95. The training process uses a batch size of 64 and a minibatch size of 32, with each PPO update performed over 4 epochs. Clipping for the policy objective is set to 0.2 to ensure stable updates. Reward normalization and observation normalization are applied throughout training to enhance learning stability.

\subsubsection{Network Architectures}
The object recognition component of the perception pipeline begins with object detection using the Segment Anything Model (SAM)~\cite{kirillov2023segment}, which generates instance-level masks for each object in the scene. These masked images are then processed by PoseCNN~\cite{xiang2018posecnn}, which utilizes a 13-layer VGG16-style convolutional backbone for feature extraction. This is followed by a semantic segmentation head with convolutional layers for per-pixel classification.

For policy learning, the PPO agent employs separate actor and critic networks, both implemented as multi-layer perceptrons (MLPs). The observation space consists of images with dimensions $2 \times 640 \times 480$ (target mask and object-level depth map, two frames stacked along the channel dimension). These inputs are first processed through a sequence of convolutional layers for feature extraction: a 32-filter convolutional layer with an $8 \times 8$ kernel and stride 4, followed by a 64-filter layer with a $4 \times 4$ kernel and stride 2, and a final 64-filter layer with a $3 \times 3$ kernel and stride 1, each followed by $\mathrm{ReLU}$ activations. The resulting feature map is flattened and passed to the actor and critic MLPs, each comprising two hidden layers with 64 nodes per layer and $\tanh$ activation functions. The actor and critic maintain independent parameter sets, enabling separate learning of the policy and value functions. Orthogonal weight initialization is applied throughout the networks, with biases initialized to zero.

%% file: appendix/nbv.tex
\subsection{Next-best-view Strategy}
\label{app:nbv}

To determine the next-best-view (NBV), our method first constructs a Truncated Signed Distance Function (TSDF) representation of the current scene, which integrates depth information from previous observations into a 3D volumetric map. This TSDF map enables the system to simulate various candidate camera viewpoints without physically moving the robot.

For each candidate viewpoint, the method predicts the grasp quality score for the target object. After evaluating all candidates, the viewpoint with the highest predicted grasp quality is selected as the next camera pose.

This NBV strategy, inspired by affordance-based NBV approaches~\cite{zhang2023affordance}, enhances perception by increasing viewpoint diversity, leading to a more comprehensive understanding of cluttered scenes and improving generalization across tasks. Unlike traditional NBV methods that may trigger viewpoint adjustments at fixed intervals, our approach selectively activates NBV only when no higher-priority actions (e.g., grasping or occlusion removal) are viable. This conditional triggering avoids unnecessary motions and ensures that NBV is used specifically to gather informative observations, enhancing task efficiency without compromising performance.

%% file: appendix/additional_exp.tex
\subsection{Additional Experimental Results}
\label{app:additional_exp}

\begin{table}[ht]
\vspace{-1em}
\centering
\caption{\textbf{Efficiency in short-horizon tasks.} XPG-RL maintains higher success rates while significantly reducing the number of actions compared to MPGNet in simpler scenes, demonstrating adaptability and avoidance of redundant motions.}
\resizebox{0.48\textwidth}{!}{
\begin{tabular}{lcccccccc}
\toprule
\multirow{2}{*}{\textbf{Method}} & \multicolumn{3}{c}{\textbf{Task success rate [\%]}} & \multicolumn{3}{c}{\textbf{Average motion number}} \\
\cmidrule(lr){2-4} \cmidrule(lr){5-7}
& 3 objects & 4 objects & 5 objects & 3 objects & 4 objects & 5 objects\\
\midrule
MPGNet & 80 & 79 & 76 & 3.02 & 3.51 & 3.84 \\
XPG-RL & 92 & 86 & 81 & 1.70 & 2.12 & 2.31 \\
\bottomrule
\end{tabular}
}
\label{tab:sim_comp_with_baseline_simple}
\vspace{-1em}
\end{table}